\documentclass[sigconf,natbib=true,anonymous=false]{acmart}
\usepackage{multirow}
\usepackage{algorithm}
\usepackage{algorithmic}
\usepackage{graphicx}
\usepackage{stfloats}
\usepackage{caption}  
\usepackage{subcaption}

\AtBeginDocument{%
  }

\setcopyright{acmlicensed}
\copyrightyear{2025}
\acmYear{2025}
\acmDOI{10.1111/123456789.123456}

\acmConference[UnderReivew]{xxxxx}
\acmISBN{978-1-1234-0234-9/18/06}

\newtheorem{definition}{Definition}
\newtheorem{assumption}{Assumption}

\acmSubmissionID{100}

\begin{document}

\title{Deep Disentangled Representation Network for Treatment Effect Estimation}


\author{Hui Meng}
\authornote{These authors contributed equally to this work and should be considered co-first authors.}
\email{congwen.mh@alibaba-inc.com}
\affiliation{%
  \institution{Alibaba Group}
  \city{Hangzhou}
  \state{Zhejiang}
  \country{China}
  }
  
\author{Keping Yang}
\authornotemark[1]
\email{shaoyao@alibaba-inc.com}
\affiliation{%
  \institution{Alibaba Group}
  \city{Hangzhou}
  \state{Zhejiang}
  \country{China}
  }

\author{Xuyu Peng}
\email{xijiu.pxy@alibaba-inc.com}
\affiliation{%
  \institution{Alibaba Group}
  \city{Hangzhou}
  \state{Zhejiang}
  \country{China}
  }

\author{Bo Zheng}
\email{bozheng@alibaba-inc.com}
\affiliation{%
 \institution{Alibaba Group}
 \city{Hangzhou}
 \state{Zhejiang}
 \country{China}
 }


\renewcommand{\shortauthors}{Hui Meng, Keping Yang et al.}

\begin{abstract}
Estimating individual-level treatment effect from observational data is a fundamental problem in causal inference and has attracted increasing attention in the fields of education, healthcare, and public policy.
In this work, we concentrate on the study of disentangled representation methods that have shown promising outcomes by decomposing observed covariates into instrumental, confounding, and adjustment factors. However, most of the previous work has primarily revolved around generative models or hard decomposition methods for
covariates, which often struggle to guarantee the attainment of precisely disentangled factors. In order to effectively model different causal relationships,
we propose a novel treatment effect estimation algorithm that incorporates a mixture of experts with multi-head attention and a linear orthogonal regularizer to softly
decompose the pre-treatment variables, and simultaneously eliminates selection bias via importance sampling re-weighting techniques.
We conduct extensive experiments on both
public semi-synthetic and real-world production datasets. The experimental results clearly demonstrate that our algorithm outperforms the state-of-the-art methods focused on individual treatment effects.
\end{abstract}




\maketitle

\section{Introduction}
Causal inference is a powerful statistical modeling tool for explanatory analysis and  plays an important role on data-driven decision making in many fields, 
such as healthcare, public policy and social marketing~\cite{RubinDonaldB1974Eceo,10.1214/09-SS057,pearl2016causal,zhanglu2019,hernan2023causal}. One fundamental problem in causal inference is treatment effect estimation from observational data, and
its key challenge is to remove the confounding bias induced by the different confounder distributions between treated and control units.
In medicine scenarios, doctors attempt to identity which medical procedure $t \in \mathcal T$ will benefit a certain patient $x$ the most, in terms of the 
treatment outcome $y \in \mathbb{R}$~\cite{Hassanpour2019}. It often requires answering counterfactual questions like ``\textit{Would this patient have lived longer, had she received an alternative treatment?}''.

One golden standard approach to learn causal effect is to perform Randomized Controlled Trials (RCTs), where the treatment is randomly assigned to individuals~\cite{10.1214/09-SS057}.
However, in many scenarios, randomized experiments are usually expensive, unethical, or even unfeasible\cite{Kohavi2011}. In light of this, we are forced to approximate causal effects from off-line datasets collected by observational studies. In such datasets, the treatment effect often depends on some covariants of the individual $x$, which causes the
problem of \textbf{selection bias}, i.e. $\displaystyle Pr(T | X) \neq Pr(T)$~\cite{imbens2015causal}. 
Consequently, it is essential to pinpoint all confounding variables and control them to make accurate predictions.
This means that unconfoundedness assumption often needs to be satisfied in the observational study to make the treatment effect identifiable.

\begin{figure}[bp]
    \centering
    \begin{subfigure}[b]{0.21\textwidth}
        \includegraphics[width=\textwidth,keepaspectratio]{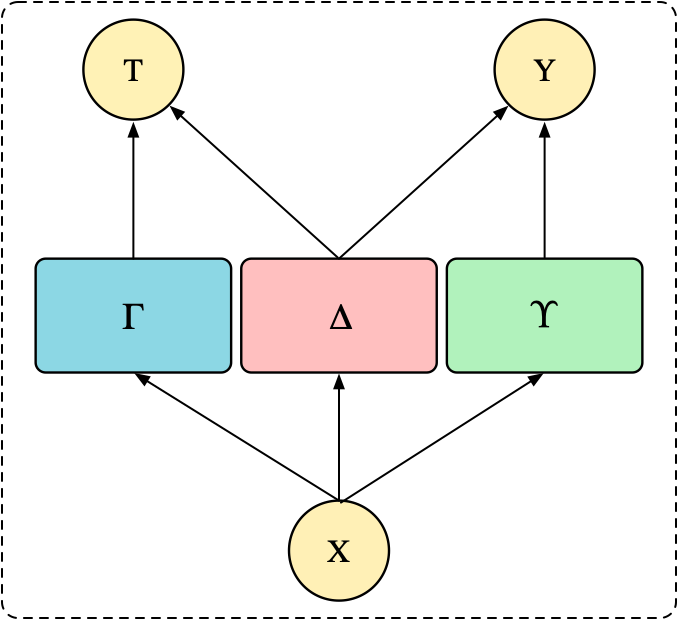}
        \captionsetup{font=scriptsize}
        \caption{The hard decomposition framework with three isolated networks.}
        \label{fig:causal4-1}
    \end{subfigure}
     \hspace{1mm}
     \begin{subfigure}[b]{0.21\textwidth}
        \includegraphics[width=\textwidth,keepaspectratio]{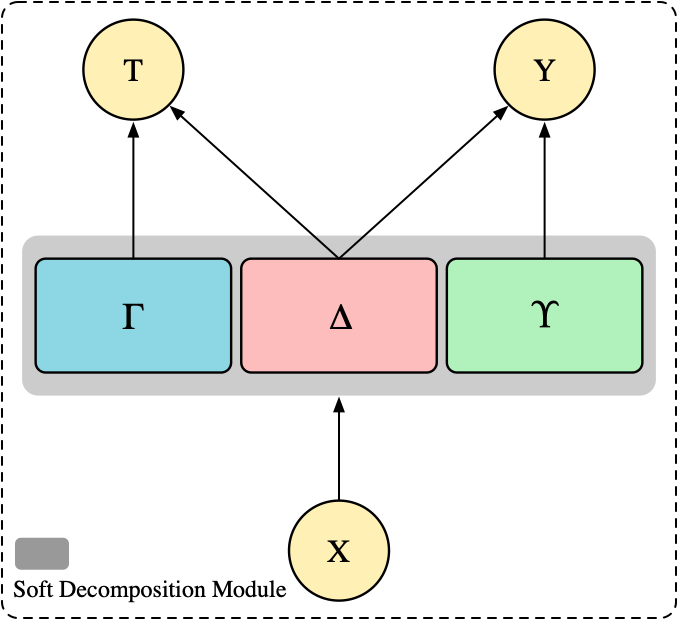}
        \captionsetup{font=scriptsize}
        \caption{The soft decomposition framework we propose.}
        \label{fig:causal4-2}
    \end{subfigure}
    \caption{The intuitive illustration of two distinct causal frameworks, where Figure~\subref{fig:causal4-1} shows the traditional hard decomposition architecture, 
    and Figure~\subref{fig:causal4-2} represents the soft decomposition approach we propose.}
    \label{fig:causaldiagram}
\end{figure}

The key to solving selection bias is identifying confounders and balancing them based on the back-door criterion~\cite{causality}.
Most of the previous methods adopt the propensity score to
re-weight units for removing confounding bias under no unmeasured confounding assumption~\cite{austin2011introduction}. 
Although these methods are gaining ground in applied work,
they require correct model specification on treatment assignment or accurate propensity score estimation. 
It is worth mentioning that variable balancing methods have also played an important role\cite{zubi2015,Hainmueller_2012}.
Nevertheless, simply balancing covariates that mixed with non-confounders would increase the bias and variance of treatment effect estimation. Recently, the disentangled representation learning methods have been proposed to roughly decompose
the pre-treatment variables into $\{ \Gamma, \Delta, \Upsilon \}$ factors, 
a.k.a, instrumental, confounding and adjustment factors~\cite{Hassanpour&Greiner2020}. 
Nonetheless, we still face a difficult problem of
identifying the instrumental, confounding and adjustment variables precisely, which is a necessary condition for identifying treatment effect.
The Deep Orthogonal Regularizer (DOR) method was proposed in~\cite{AnpengWu2022}, aims to obtain accurate disentangled representations. 
Unfortunately, regularizing the weight matrix may lead to a challenging decomposition problem of the covariates, as shown in Figure \ref{fig:causal4-1}.



In this work, we propose an easy-to-use and robust causal learning model named the Deep Disentangled Representation Network to address the challenges
associated with disentangled representation,
which primarily includes a novel multi-task learning framework, a linear orthogonal regularizer, and balancing constraints. 
The main contributions of this paper are summarized as follows:

\begin{itemize}

\item We propose an innovative multi-task learning structure that combines a mixture of experts with multi-head attention, which enhances information acquisition and factor identification performance compared to using three isolated networks or a single bottom-shared architecture.
\item We choose to softly decompose the pre-treatment factors in the latent representation space, rather than rigidly separate them at the input layer, due to the higher fault tolerance and robustness of this method, as illustrated in Figure \ref{fig:causal4-2}.
\item We introduce a novel linear orthogonal regularizer and multiple interpretable representation balancing constraints for CounterFactual Regression (CFR), which could obtain ideally independent disentangled representations.

\item We conduct extensive experiments on both public benchmark datasets and real-world production datasets, with the results indicating that our approach achieves a better performance of individual treatment effect estimation in observational studies. 
\end{itemize}

The rest of the article is organized as follows. Section 2 reviews the related work. Section 3 gives problem setting and assumptions. Section 4 introduces our proposed 
method. Experimental results and analyses are reported in Section 5. The conclusion and future outlook of the paper are presented in Section 6. 
Finally, Section 7 is the ethical statement.

\section{Related Work}

Treatment effect estimation in observational studies faces the challenge of reducing selection bias, which has drawn the attention of many researchers due to its broad applicability across different fields.
The propensity score which was defined by~\cite{RosenbaumandRubin1983}  is one of the mainstream approaches to eliminate 
confounding bias. Without loss of generality,~\cite{austin2011introduction}  pointed out that there are four categories of propensity score based methods used for removing 
effects of confounding, including matching, stratification, inverse propensity weighting and covariate adjustment using the propensity score \cite{Schafer&Kang2009,Joffe2004,Lunceford2004,austin2011introduction,pearl2016causal,rubin1997}. 

\cite{Johansson&Shalit2016} proposed a representation learning method for confounder balancing by minimizing
distribution between treated and control groups in embedding space.~\cite{shi2019adapting} induced the Dragonnet neural architecture, which provides an end-to-end 
procedure for predicting propensity score and conditional outcome from covariates and treatment information. Based on the advances,~\cite{Hassanpour2019}
proposed a context-aware re-weighting scheme based on importance sampling technique, on top of a representation learning module, to alleviate the problem of selection bias 
in causal effect estimation. In addition, \cite{Kuang2019,AnpengWu2022} pointed out the necessity of confounder separation or selection for causal inference, 
due to the fact that the control of some non-confounders would generate additional bias and amplify the variance, especially in high-dimensional settings.

There has been a notable rise in attention toward disentangled representation learning in the latest research trends.
\cite{KuangKun2017} proposed a data-driven disentangled representation algorithm to automatically separate confounders and adjustment variables. 
The limitation of this method is that they ignored differentiation between instrumental and confounding factors, apparently leading to imprecise confounder identification. \cite{Hassanpour&Greiner2020} argued that explicit identification of the underlying factors 
$\displaystyle \{ \Gamma, \Delta, \Upsilon \}$ in observational datasets offers great insight to handle selection bias, and then achieve
better performance in terms of estimating individual treatment effect. Nevertheless, this method can not guarantee accurately learning disentangled representations.
CEVAE employed variational autoencoder to learn confounders from observed covariates, but it does not consider the existence of non-confounders~\cite{louizos2017causal}.
Building upon this work,~\cite{zhang2021treatment}  then utilized a similar variational inference architecture to simultaneously learn and disentangle the latent factors from observed variables into three disjoint subsets.
However, this work still suffers from the aforementioned limitations, Furthermore, the generative model which might greatly increase model training complexity and time overhead.

In order to obtain independent disentangled representations, \cite{cheng2022learning} used a basic multi-task learning framework to share information 
when learning the latent factors and incorporated mutual information minimization learning criteria to ensure the independence of these factors.
Nevertheless, this method utilized a variational distribution $\displaystyle q_{\theta}(y|x)$ to approximate the conditional distribution $p(y|x)$ of 
Contrastive Log-ratio Upper Bound, which was defined as:
$\mathrm{I_{vCLUB}} (x, y) = \mathbb E_{p(x,y)} [\mathrm{log} q_{\theta}(y|x)] - \mathbb E_{p(x)} \mathbb E_{p(y)} [\mathrm{log} q_{\theta}(y|x)]$.
There is no doubt that variational approximation error will negatively impact the model learning accuracy. \cite{AnpengWu2022} employed a deep orthogonal regularizer
among three representation networks for decomposing the pre-treatment variables $X$ into instrumental, confounding and adjustment factors $\{I, C, A\}$. The main weakness of this method is the computational complexity is too high. We take the representation network $I(X)$ for instrumental factor as an example. The time complexity of $I(X)$ may be
expressed as $\mathcal {O} (HMD)$, where $M$ is input features, $H$ is hidden layer dimension, and $D$ defines the final returned output. It is worth noting that all the aforementioned methods perform regularization within three independent networks, which makes it challenging to ensure the attainment of accurate disentangled representations. 

\begin{figure*}[htbp]
\centering
\includegraphics[width=0.845\textwidth,keepaspectratio]{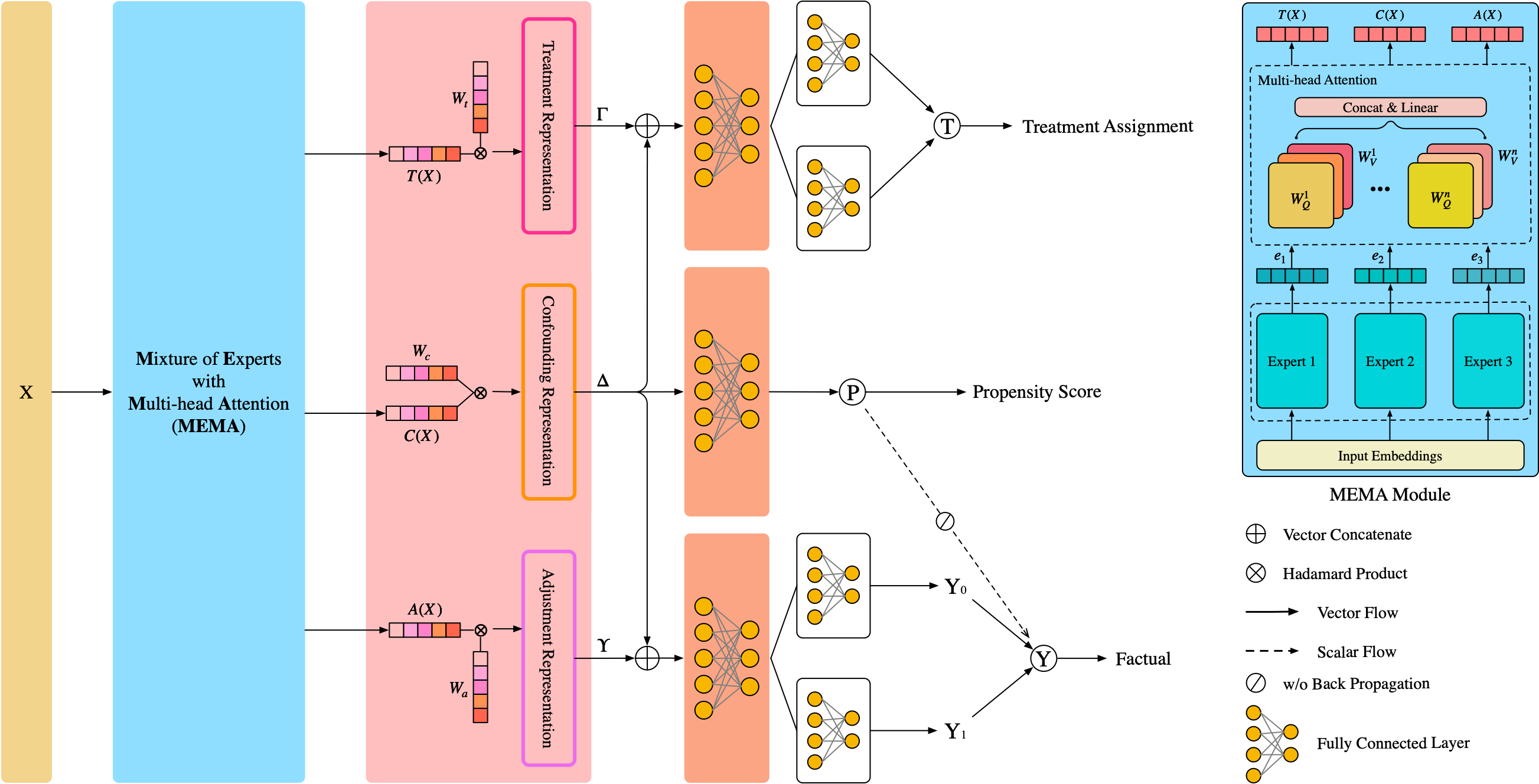}
\caption{The overall network architecture of our proposed model. This approach takes user profiles, behavior sequences, and real-time contextual information (if available) as input covariates.
It firstly embeds these covariates into low-dimensional vector representations. In order to improve the capacity for information extraction and factor identification,  
MEMA layer is utilized to learn the representations of treatment, confounder, and adjustment. Then, the linear orthogonal regularizer ensures that precise soft-separation results can be obtained. Finally, the concatenated $\Gamma$ and $\Delta$ flows into the treatment assignment task, $\Delta$ and $\Upsilon$ are concatenated and fed into the regression networks, while $\Delta$ is independently used to predict the propensity score.}
\label{fig:figure09}
\end{figure*}

With these improvements, we propose a linear orthogonal regularizer that requires only a small amount of computation resources to obtain independent disentangled representations. In addition, we introduce a novel mixture of experts with multi-head self-attention structure that not only ensures $\{ \Gamma, \Delta, \Upsilon \}$ originate
from the same representational space but also improves information acquisition and factor identification ability, as shown in Figure \ref{fig:figure09}.

\section{Problem Setting and Assumptions}

In this paper, we aim to estimate the individual treatment effect from observational data. 
Given the observational dataset  $\displaystyle D= \{  x_{i}, t_{i}, y_{i}^{t_{i}} (x_i, t_i) \}_{i=1}^{n}$, where $n$ is the number of data samples. For each
unit $i$, we observe its context characteristics $x_{i} \in \mathcal{X}$, potential intervention $t_{i} \in \mathcal T$ from a set of treatment options 
(e.g., for binary treatment $t \in \{0, 1\}$), and the corresponding factual outcome $y_{i}^{t_{i}} (x_i, t_i) \in \mathcal Y$ as a result of choosing treatment $t_{i}$.
Noting that due to the counterfactual problem, we cannot observe the potential outcome for treated units.
Mathematically, we define our goal in this work is to learn a 
function $\mathcal F: \mathcal X \times \mathcal T \rightarrow \mathcal Y$ to predict the potential outcomes and then estimate the Individual Treatment Effect (ITE)
and the Average Treatment Effect (ATE). 

In the context of our research, we focus primarily on scenarios involving binary treatment. Therefore, the definition of ITE is as follows:

\begin{definition}[The Individual Treatment Effect] 
\begin{equation} 
\resizebox{.85 \linewidth}{!} {$
ITE(x) := \mathbb E [\mathbf y|\mathbf X = x, do(\mathbf t=1)] - \mathbb E [\mathbf y|\mathbf X=x, do(\mathbf t=0)]
$}
\end{equation}
\end{definition}

With the knowledge of ITE, we can readily assess the ATE value, which is defined as follows:

\begin{definition}[The Average Treatment Effect]
\begin{equation} 
ATE := \mathbb E[ITE(x)] = \mathbb E[y^1-y^0]
\end{equation}
\end{definition}

The following fundamental assumptions~\cite{imbens2015causal} need to be satisfied in individual causal effect estimation:
\begin{assumption}[Stable Unit Treatment Value]
The distribution of the potential outcome of one unit is assumed to be independent of the treatment assignment of another unit.
\end{assumption}

\begin{assumption}[Unconfoundedness]
The treatment assignment mechanism is independent of the potential outcome when conditioning on the observed variables.
Formally expressed as $T\perp(Y^0, Y^1) | X$.
\end{assumption}

\begin{assumption}[Overlap]
Every individual x has non-zero chance of being assigned to any treatment arm, i.e. $\displaystyle 0 < Pr(t=1|x) < 1$.
\end{assumption}



\section{The Proposed Method}

In this section, we first illustrate the overall architecture of the proposed method. Then we detail each part of DDRN, 
including multi-task disentangled representation learning, known as the Mixture of Experts with Multi-head Attention (MEMA),
and the various components of objective function. Finally, we introduce the algorithm pseudocode of our proposed method.

\subsection{Architecture Overview}

The overall architecture of our proposed method is illustrated in Figure \ref{fig:figure09}, which primarily consists of three components: 
(1) a novel mixture of experts with multi-head attention multi-task learning structure, which enhances information acquisition and factor identification capabilities.
(2) an ingenious linear orthogonal regularizer to ensue the attainment of precise disentangle representation results. Especially, compared to previous approaches, our method achieves soft decomposition of the
$\displaystyle \{ \Gamma, \Delta, \Upsilon \}$ factors in a high-dimensional latent space, rather than roughly segmenting the pre-treatment variables at the input layer.
(3) two regression networks for potential outcomes prediction and two logistic networks to model the logging treatment policy and sample weights. 
We therefore try to minimize the following objective function:
\begin{equation}
\resizebox{.905 \linewidth}{!} {$
\mathcal L = \mathcal \omega \cdot \mathcal L_{pred} + \alpha \cdot \mathcal L_{treat} +\beta \cdot \mathcal L_{disc} + \zeta \cdot \mathcal L_{lor} 
+\eta \cdot \mathcal L_{isw} +  \mathfrak {Reg}
$}
\end{equation}
where $\mathcal L_{pred}$ is the prediction loss for observed outcomes weighted by importance weighting function $\mathcal \omega(\cdot)$.
$\mathcal L_{treat}$ represents the treatment assignment loss,
$\mathcal L_{disc}$ denotes the imbalance loss,
$\mathcal L_{lor}$ and $\mathcal L_{isw}$ refer to the disentanglement loss and
importance re-weighting loss, respectively.
 $\alpha$, $\beta$, $\zeta$, and $\eta$ are given hyperparameters that control the weights of the aforementioned loss terms. 
 Furthermore, $\mathfrak {Reg}$ refers to the $L_2$ regularization term on the model parameters.

\subsection{Mixture of Experts with Multi-head Attention}

We propose a novel multi-task learning framework called Mixture of Experts with Multi-head Attention (MEMA) see Figure~\ref{fig:figure09}, which aims to precisely capture the differences between tasks. Compared to the classic MMoE method~\cite{MMoE},
the key advantage lies in the multi-head self-attention mechanism, which guarantees that the disentangled representations derive from the same hidden space, 
thereby enabling more accurate soft decomposition results.
More specifically, the output of task $k$ can be represented as follows:
\begin{equation}
f^{k}(x) = \mathrm{FFN}(\mathrm{Concat}(\mathrm{head_1},\cdots,\mathrm{head_n}))^k
\end{equation}
where $\mathrm{head}$ represents scaled dot-product attention, $\mathrm{Concat}$ means concatenation, $\mathrm{FFN}$ refers to feed-forward network,.


%

In terms of implementation, the expert networks consist of canonical multi-layer perceptrons with GELU activations, 
which transform the pre-treatment variables into latent vectors, as indicated by $E = \{e_1(x), \cdots, e_m(x)\}$,
where $m$ represents the number of expert networks. Immediately afterwards, add either relative or absolute positional encodings to the input embeddings $E$ at the bottom of the encoder stacks.
We adopt multi-head attention that is identical to that used in the Transformer,
which allows the model to jointly attend to
information from different representation subspaces, whereas a single attention head tends to suppress this capability~\cite{vaswani2023attention}. 
The $i$-th head attention operation can be expressed as follows:
\begin{equation}
\mathrm{head_i} = \mathrm{Attention}(QW_i^Q, KW_i^K , VW_i^V)
\end{equation}
where the projections are parameter matrices $W_i^Q \in \mathbb{R}^{d_m \times d_k}$, $W_i^K \in \mathbb{R}^{d_m \times d_k}$ and $W_i^V \in \mathbb{R}^{d_m \times d_v}$. It is worth mentioning that under orthogonal constraints, this type of self-attention mechanism can facilitate information exchange, allowing different tokens to focus on learning distinct representations.

\subsection{Factural Loss}
Considering the binary treatment, we train two regression networks $h^{0}$ and $h^{1}$, one for each treatment arm. Concretely, we concat $\Upsilon$ and $\Delta$
latent representations as input to the regression networks.
\begin{equation}
\mathcal L_{pred} = \mathcal L [y, h^{t}(\Delta(x), \Upsilon(x))]
\end{equation}
Notably, since observational data never includes the counterfactual outcomes for any training instances, the regression loss $\mathcal L_{pred}$ can only be calculated on the
factual outcomes. We ensure that the union of the learned representations $\Delta$ and $\Upsilon$ retain enough information needed for accurate estimation of
the observed outcomes by minimizing the factural loss.

\subsection{Treatment Loss}
In order to prevent the information of instrumental variables from being embedded into other representation factors, 
we introduce a logistic regression network $\pi$ to model the treatment assignments. Therefore, treatment loss function formula as follows:
\begin{equation}
\mathcal L_{treat} = \mathcal L (t_{i}, \pi(\psi(x_{i})))
\end{equation}
where  $\psi$ is the concatenation of latent representations $\Gamma$ and $\Delta$. By minimizing binary cross entropy loss $\mathcal L_{treat}$ enforces learning latent
factors $\Delta$ and $\Gamma$ in a way that allows $\pi$ to predict treatment assignments.


\subsection{Imbalance Loss}
According to the causal diagram mentioned in~\cite{Hassanpour&Greiner2020}, adjustment factor $\Upsilon$ should be independent of treatment assignments, namely $\Upsilon \perp \mathrm{T}$.
In this work, we use Maximum Mean Discrepancy (MMD) approach proposed by \cite{gretton12a} to calculate dissimilarity between the two conditional distributions of $Y$ given $t=0$ versus $t=1$.
As an alternative, the Wasserstein distance (WASS) approach \cite{arjovsky2017wasserstein} can also be employed for this purpose.
Therefore, the discrepancy of $\Upsilon$ representation distribution between different treatment arms is:
\begin{equation}
\mathcal{L}_{disc(\Upsilon \perp \mathrm{T})} = \mathrm{disc} (\{ \Upsilon(x_{i})\}_{i:t_{i}=0}, \{ \Upsilon(x_{i}) \}_{i:t_{i}=1})
\end{equation}
Under the unconfounderness assumption, the instrumental factor $\Gamma$ would be  independent of the outcome. Refer to the formula above, we can obtain the discrepancy of $\Gamma$ representation distribution between different  outcomes $Y$. Without loss of generality, we assume the outcome 
variable is binary $y_{i} \in \{ 0, 1\}$. The formula as follows:
\begin{equation}
\mathcal{L}_{disc(\Gamma \perp \mathrm{Y})} = \mathrm{disc} (\{ \Gamma(x_{i})\}_{i:y_i=0}, \{ \Gamma(x_{i}) \}_{i:y_{i}=1})
\end{equation}
For continuous or multi-valued outcome, we can approximately achieve the conditional independence by minimizing the Mutual Information (MI) between $\Gamma$ and $Y$.
To sum up, the imbalance loss can be expressed as:
\begin{equation}
\mathcal{L}_{disc} = \mathcal{L}_{disc(\Upsilon \perp \mathrm{T})}+\mathcal{L}_{disc(\Gamma \perp \mathrm{Y})}
\end{equation}


\subsection{Importance Sampling Weighting Loss}
We refer to the approach of context-aware importance sampling weighting shown in Equation \ref{Eq.1} that employs $\Delta$ latent factor instead of $\{\Gamma, \Delta\}$ to calculate propensity score, as described in \cite{Hassanpour2019,Hassanpour&Greiner2020}.
\begin{equation}
\omega_{i} = 1 + \frac{\mathrm{Pr}(\Delta_{i} | \neg t_{i})}{\mathrm{Pr}(\Delta_{i} | t_{i})} = 1 + \frac{\mathrm{Pr}(t_{i})}{1 - \mathrm{Pr}(t_{i})} \cdot \frac{1-g(t_{i} | \Delta_{i})} {g (t_{i} | \Delta_{i})} \label{Eq.1}
\end{equation}
The reason lies in the factual loss $\mathcal L_{pred}$ is only sensitive to the latent factors $\Delta$ and $\Upsilon$,
and not to $\Gamma$, re-weighting the factual loss $\mathcal L_{pred}$ by introducing $\Gamma$ would result in emphasizing the wrong instances\cite{SHIMODAIRA2000227}. Therefore,
the formula of importance weighting loss is defined as: 
\begin{equation}
\mathcal{L}_{iw}=\frac{1}{N}\sum_{i=1}^{N}-\mathrm{log}[g(\Delta(x))]
\end{equation}
where $g(\cdot)$ is parametrized by the Logistic Regression (LR) model, shown as follows:
 \begin{equation}
 g(t | \Delta(x))= \frac{1}{1+e^{-(2t-1) (\Delta(x) \cdot W+b)}}
 \end{equation}
 and parameters $W$ and $b$ can be learned by minimizing $\mathcal{L}_{iw}$.

\subsection{Linear Orthogonal Regularizer}
Although we have introduced MEMA structure to learn disentangled latent representations, deep neural network tends to overfit the training data, especially small dataset, and
then leads to unclean disentanglement. Inspired by the orthogonal regularizer methods for variable decomposition\cite{orthreg2018}, we employ a simple linear orthogonal regularizer
to learn precise and clean disentangled representations. Assuming that we randomly initialize three vectors $\{ \mathbf W_{t}, \mathbf W_{c}, \mathbf W_{a}\}$,
and perform element-wise product with the representations $\{\mathbf T(X), \mathbf C(X), \mathbf A(X)\}$ obtained from MEMA, respectively. Taking the treatment representation $\Gamma$ as an example, the element-wise product of $\mathbf W_{t}$ and $\mathbf T(X)$ can be expressed as:
\begin{equation}
 \Gamma = \mathbf W_{t} \otimes \mathbf T(X) =  (w_1 \times t_{1}(x), \ldots,  w_n \times t_{n}(x))
 \end{equation}
 where $n$ refers to the representation dimension. 

By minimizing the distances on each pair of them $\{ \mathbf W_{t}, \mathbf W_{c}, \mathbf W_{a}\}$, we ensure that the learned latent factors
maintain orthogonality. The loss term is as follows:
\begin{equation}
\mathcal L_{lor} = \mathbf W_{t} \cdot \mathbf W_{c} + \mathbf W_{c} \cdot \mathbf W_{a} + \mathbf W_{t} \cdot \mathbf W_{a}
\end{equation}
From the above formula, we can see that the time complexity may be expressed as $\mathcal {O} (H)$, where $H$ is hidden layer dimension, that is output dimensions
of expert network. Obviously, we have greatly reduced the computational complexity compared to \cite{AnpengWu2022}.

\subsection{Algorithm Pseudocode}
To more clearly demonstrate the computational process and implementation details of the algorithm, we have described the pseudocode in Algorithm \ref{alg:algorithm} and 
open-sourced the core code of the DDRN algorithm on GitHub\footnote{\url{https://github.com/luckyradiant/sigir960}}.

\section{Experiments}
Evaluating causal inference models is always challenging because we usually lack ground truth for the counterfactual outcomes. A common solution is to synthesize datasets, where the outcomes of all possible treatments are available. Following that, some entries are discarded in order to create a biased observational dataset.
In this work, we conduct the experiments on both public benchmark datasets and a real-world production dataset.

\begin{algorithm}[htp]
    \caption{DDRN: Disentangled Representation Network for Treatment Effect Estimation}
    \label{alg:algorithm}
    \begin{algorithmic}[1] 
    	\STATE \textbf{Input:} observational dataset: $\mathcal D = \{x_{i}, t_{i}, y_{i}^{t_{i}}\}_{i=1}^{N}$
	\STATE \textbf{Parameter}: Set learning rate $lr$, mini-batch size $m$, scaling parameters \{$\alpha$, $\beta$, $\zeta$, $\eta$, $\lambda$\}, and limit on the total number of iterations $\mathcal I$. Initialize networks $\mathrm{MEMA}$, $h$, $\pi$, $g$ and Linear Orthography Regularizer (LOR) with random weights. 
	\STATE \textbf{Output}: $\hat{y}_{0}$ and $\hat{y}_{1}$
        \FOR {$i=1$ to $\mathcal I$}
        \STATE Sample mini-batch of $m$ records $\{x_{i}, t_{i}, y_{i}\}$ from $\mathcal D$
        \STATE $\mathrm{LOR}(\mathrm{MEMA}(x_{i}, t_{i}, y_{i})) \rightarrow \{ \Gamma(\cdot), \Delta(\cdot), \Upsilon(\cdot) \}$
        \STATE $h^{t}(\mathrm{Concat}(\Delta(x), \Upsilon(x))) |_{t \in \{0,1\}} \rightarrow  \hat{y}^{0}|_{t=0},  \hat{y}^{1}|_{t=1}$
        \STATE $\pi(\mathrm{Concat}(\Delta(x), \Gamma(x))) \rightarrow  \hat{t}$
        \STATE $\omega(g(\Delta(x))) \rightarrow \hat{\omega}$, where $g(\cdot)=\frac{1}{1+e^{-(2t-1) \cdot z}}$
        \STATE Calculate $\mathcal{L}_{disc}$ and $\mathcal{L}_{lor}$ loss, respectively
        \STATE Update DDRN by minimizing the total loss: \\
        $\mathcal L = \mathcal \omega \cdot \mathcal L_{pred} + \alpha \cdot \mathcal L_{treat} +\beta \cdot \mathcal L_{disc} + \zeta \cdot \mathcal L_{lor} + \eta \cdot \mathcal L_{iw}  + \mathfrak {Reg}$
        \ENDFOR
        \STATE \textbf{return} solution
    \end{algorithmic}
\end{algorithm}

\subsection{Datasets}
\textbf{Benchmark \uppercase\expandafter{\romannumeral1}: Infant Health and Development Program}\footnotemark{}

The original Randomized Controlled Trials (RCTs) data of the Infant Health and Development Program (IHDP) aims at evaluating the effect of specialist home visits on the future cognitive test scores of premature infants. The dataset comprises 747 units (139 treated, 608 control) with 25 pre-treatment variables related
to the children and their mothers. We report the estimation errors on the same benchmark (100 realizations of the outcomes with 63/27/10 proportion of train/validation/test splits) provided by and used in ~\cite{Johansson&Shalit2016,shalit2017estimating}.

\noindent \textbf{Benchmark \uppercase\expandafter{\romannumeral2}: Jobs}\footnotemark[\value{footnote}]
\footnotetext{\url{https://www.fredjo.com/}}

The Jobs dataset created by~\cite{LaLonde1986} is a widely used benchmark in the causal inference community, based on the randomized controlled trials. 
The dataset aims to estimate the effect of job training programs on employment status. Jobs contains 17 variables, such as age, education level, etc.
Following~\cite{Smith&Todd2005}, we use LaLonde’s data (297 treated, 425 control) and the PSID comparison group (2490 control) to carry out our experiment. We randomly 
split the data of 3212 samples into train/validation/test with a 56/24/20 ratio (10 realizations).

\noindent \textbf{Benchmark \uppercase\expandafter{\romannumeral3}: 2016 Atlantic Causal Inference Challenge}\footnote{\url{https://github.com/vdorie/aciccomp/}}

The 2016 Atlantic Causal Inference Challenge (ACIC 2016)~\cite{VincentDorie2019} contains 77 different settings of benchmark datasets that are designed to test causal 
inference algorithms under a diverse range of real-world scenarios. Concretely, the dataset contains 4802 observations and 58 variables. The treatment and outcome
variables are generated using different data generating procedures for the 77 settings, providing benchmarks for a wide range of treatment effects estimation scenarios.
~\cite{zhang2021treatment}.

\noindent \textbf{Real-world Production Dataset: Message Pop-up Dataset}

The Message Pop-up Dataset (MPD) is collected from online marketing policy experiments, a particular randomized trial procedure where a random part of the population is prevented from being targeted by pop-up promotion. 
The dataset is composed of 10 million records, with each record corresponding to an individual user.
The treatment variable is defined as whether a pop-up message was displayed on the homepage of the \textbf{Xxxxxx} application\footnote{The Xxxxxxxx Xxxxx's e-commerce platform}, and a positive outcome denotes that the user converted on the same day.
We randomly split the data into train, validation, and test sets with a 70\%, 10\%, and 20\% ratio, respectively.
In order to satisfy the unconfoundedness assumption, we introduce sufficient confounders based on our prior knowledge.

\subsection{Performance Metrics}

\begin{figure}[tbp]
\centering
\includegraphics[width=0.42\textwidth,keepaspectratio]{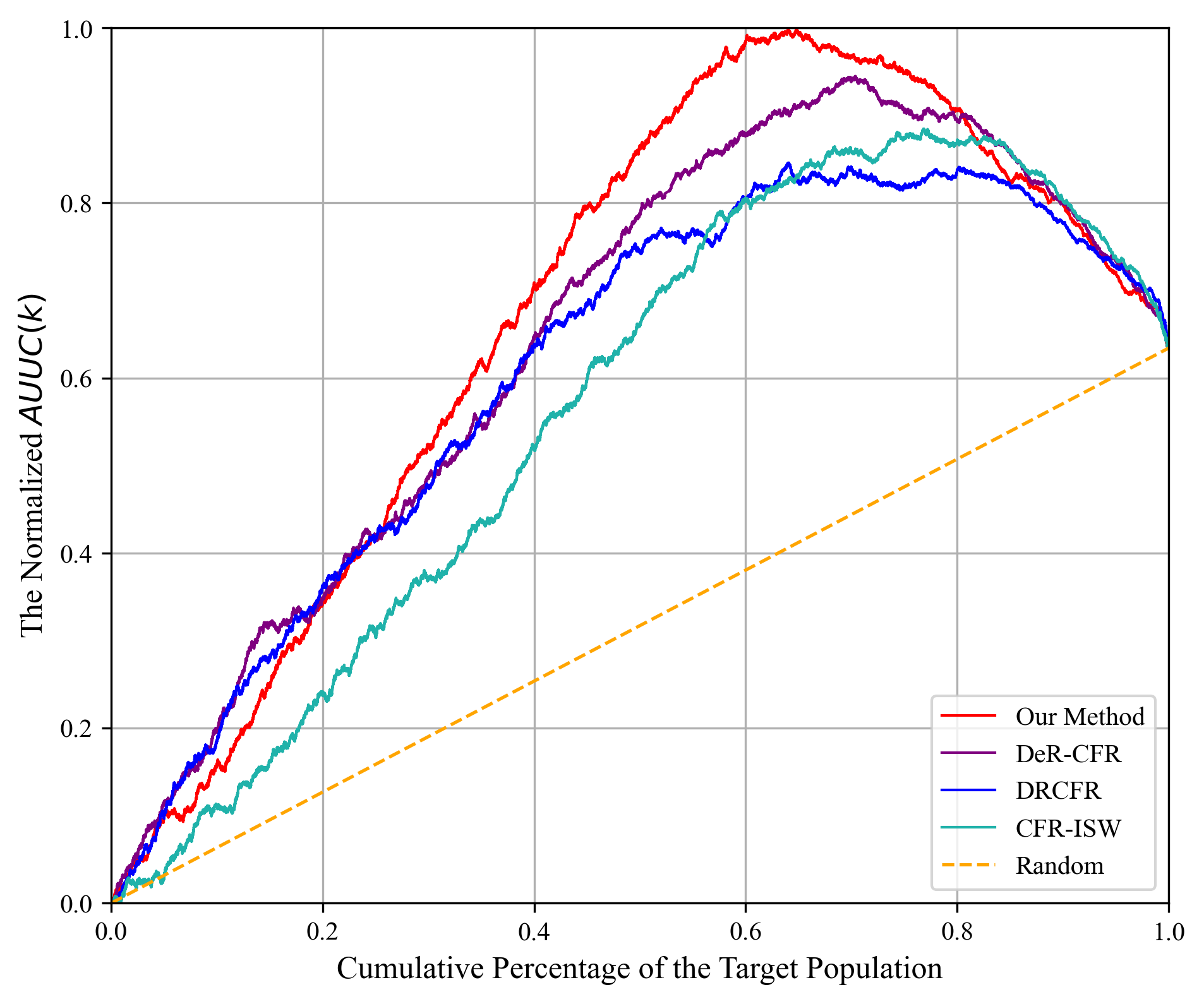}
\caption{Uplift curves for different models on Message Pop-up Dataset, where x-axis denotes the proportion $k$ of the test dataset, y-axis denotes the normalized $AUUC_{\pi}(k)$
value by dividing the maximum value of $AUUC_{\pi}(k)$.}
\label{fig:auuc_ddrn}
\end{figure}

On IHDP and ACIC 2016 datasets, which include both factual and counterfactual outcomes, we adopt the Precision in Estimation of Heterogeneous Effect as the individual-level performance metric\cite{hill21011}, the formula is as follows:
\begin{equation}
\mathrm{PEHE} = (\frac{1}{N} \sum_{i=1}^{N} (\hat{\tau}(\mathbf x_{i})-\tau(\mathbf x_{i}))^2)^{\frac{1}{2}}
\end{equation}
where $\tau(\mathbf{x}_{i})$ indicates the actual effect for subjects
with observed variables $\mathbf x_{i}$. For the population causal effect we report the absolute error on the Average Treatment Effect, the operation can be stated as follows:
\begin{equation}
\epsilon_{ATE} = |ATE - \widehat{ATE}|
\end{equation}
where $ATE = \mathbb E(y^1) - \mathbb E(y^0)$, and $\widehat{ATE}$ is calculated based on the estimated outcomes.

On Jobs dataset, there is no ground truth 	for counterfactual outcomes, so the policy risk is adopted, which is defined as:
\begin{align}
\mathcal R_{pol} = & 1 - \mathbb E[y^1 | \pi_{f}(x)=1, t=1] \cdot \mathcal P(\pi_{f}(x) =1)  \nonumber \\ - & E[y^0 | \pi_{f}(x)=0, t=0] \cdot \mathcal P(\pi_{f}(x) =0)
\end{align}
where $\pi_{f}(x) =1$ if $\hat{y}^{1} - \hat{y}^{0} > 0$ and $\pi_{f}(x) =0$, otherwise. The policy risk measures the expected loss if the treatment is taken according to the ITE estimation
\cite{shalit2017estimating}. For both PEHE and policy risk, the smaller value is, the better the performance.

In real-world scenarios, enhancing incremental value based on uplift models is an important research direction. 
As we cannot obtain factual and counterfactual outcomes simultaneously in the real-world datasets, the mainstream metrics for offline evaluation include the Area Under the Uplift 
Curve (AUUC) and the Qini coefficient \cite{hal02515860,learn2rankum}.

AUUC is obtained by subtracting the respective area under lift curve~\cite{upliftcurves}, 
which represents the proportion of positive outcomes as a function of the percentage of the individuals selected, as follows: 
\begin{equation}
AUUC_{\pi}(k) = \sum_{i=1}^{k}(R_{\pi}^{T}(i) - R_{\pi}^{C}(i)) - \frac{k}{2}(\bar{R}^{T}(k) - \bar{R}^{C}(k))
\end{equation}
where $R_{\pi}^{T}(k)$ and $R_{\pi}^{C}(k)$ denote the numbers of positive outcomes in the treatment and control groups respectively,
$\bar{R}^{T}(k)$ and $\bar{R}^{C}(k)$ are the numbers of positive outcomes assuming a uniform distribution of positives. 
The total AUUC is then obtained by cumulative summation:
\begin{equation}
AUUC_{\pi} = \int_{0}^{1} AUUC_{\pi}(\rho)\, d\rho \approx \frac{1}{n} \sum_{k=1}^{n} AUUC_{\pi}(k)\, dk
\end{equation}
A higher AUUC signifies better separation between treatment and control group outcomes.

Qini coefficient is a generalization of the Gini coefficient for the uplift prediction problem. The key difference compared to AUUC is that Qini corrects uplifts of selected 
individuals with respect to the number of individuals in treatment/control using the factor $N_{\pi}^{T}(k) / N_{\pi}^{C}(k)$,
Therefore, we can define the Qini coefficient for binary outcomes as ratio of the actual uplift gains curve above the diagonal to that of the optimum Qini curve:
\begin{equation}
Q_{\pi} = \int_{0}^{1} Q_{\pi}(\rho)\,d\rho / \int_{0}^{1} Q_{\pi^{*}}(\rho)\,d\rho
\end{equation}
where $\pi^{*}$ refers to the optimal ordering. We use a standard
python package scikit-uplift to compute these metrics\footnote{\url{https://www.uplift-modeling.com/en/latest/}}.

\begin{table*}[t]
\footnotesize
\renewcommand{\arraystretch}{1.1}
\setlength{\tabcolsep}{9pt}
\caption{The results (mean±std) of treatment effect estimation on benchmark datasets.}
\label{exp-results}
\begin{center}
\small
\resizebox{ 0.90\textwidth}{!}{
\begin{tabular}{ccccccccccccccc}
\toprule   
\multicolumn{15}{c}{\bf Within-sample} \\
\toprule
\multicolumn{3}{c}{\bf Datasets}& \multicolumn{4}{c}{\bf IHDP}& \multicolumn{4}{c}{\bf ACIC 2016}& \multicolumn{4}{c}{\bf Jobs} \\
\cmidrule(r){1-3} \cmidrule(lr){4-7} \cmidrule(l){8-11} \cmidrule(l){12-15} 
\multicolumn{3}{c}{Methods}& \multicolumn{2}{c}{PEHE}& \multicolumn{2}{c}{$\epsilon_{\mathrm{ATE}}$}& \multicolumn{2}{c}{PEHE}& \multicolumn{2}{c}{$\epsilon_{\mathrm{ATE}}$}& \multicolumn{2}{c}{$\mathcal R_{pol}$}& \multicolumn{2}{c}{$\epsilon_{\mathrm{ATT}}$} \\
\midrule[0.5pt]
\multicolumn{3}{c}{CFR-ISW}& \multicolumn{2}{c}{$0.598 \pm 0.028$}& \multicolumn{2}{c}{$0.210 \pm 0.028$} & \multicolumn{2}{c}{$2.235 \pm 0.170$}& \multicolumn{2}{c}{$0.250 \pm 0.041$} & \multicolumn{2}{c}{$0.189 \pm 0.006$}& \multicolumn{2}{c}{$\bf{0.041} \pm 0.017$} \\
\multicolumn{3}{c}{DRCFR}& \multicolumn{2}{c}{$0.657 \pm 0.028$}& \multicolumn{2}{c}{$0.240 \pm 0.032$} & \multicolumn{2}{c}{$2.440 \pm 0.200$}& \multicolumn{2}{c}{$0.301 \pm 0.052$} & \multicolumn{2}{c}{$0.199 \pm 0.006$}& \multicolumn{2}{c}{$0.064 \pm 0.026$} \\
\multicolumn{3}{c}{DeR-CFR}& \multicolumn{2}{c}{$0.444 \pm 0.020$}& \multicolumn{2}{c}{$0.130 \pm 0.020$} & \multicolumn{2}{c}{$1.643 \pm 0.104$}& \multicolumn{2}{c}{$0.194 \pm 0.044$} & \multicolumn{2}{c}{$0.187 \pm 0.037$}& \multicolumn{2}{c}{$0.053 \pm 0.084$} \\
\multicolumn{3}{c}{Our Method}& \multicolumn{2}{c}{$\bf{0.363} \pm 0.024$}& \multicolumn{2}{c}{$\bf{0.090} \pm 0.027$} & \multicolumn{2}{c}{$\bf{1.074} \pm 0.095$}& \multicolumn{2}{c}{$\bf{0.126} \pm 0.036$} & \multicolumn{2}{c}{$\bf{0.162} \pm 0.024$}& \multicolumn{2}{c}{$0.045 \pm 0.037$} \\
\bottomrule
\toprule
\multicolumn{15}{c}{\bf Out-of-sample} \\
\toprule
\multicolumn{3}{c}{\bf Datasets}& \multicolumn{4}{c}{\bf IHDP}& \multicolumn{4}{c}{\bf ACIC 2016}& \multicolumn{4}{c}{\bf Jobs} \\
\cmidrule(r){1-3} \cmidrule(lr){4-7} \cmidrule(l){8-11} \cmidrule(l){12-15}
\multicolumn{3}{c}{Methods}& \multicolumn{2}{c}{PEHE}& \multicolumn{2}{c}{$\epsilon_{\mathrm{ATE}}$}& \multicolumn{2}{c}{PEHE}& \multicolumn{2}{c}{$\epsilon_{\mathrm{ATE}}$}& \multicolumn{2}{c}{$\mathcal R_{pol}$}& \multicolumn{2}{c}{$\epsilon_{\mathrm{ATT}}$} \\
\midrule[0.5pt]
\multicolumn{3}{c}{CFR-ISW}& \multicolumn{2}{c}{$0.715 \pm 0.102$}& \multicolumn{2}{c}{$0.218 \pm 0.031$}& \multicolumn{2}{c}{$2.404\pm 0.185$}& \multicolumn{2}{c}{$0.289 \pm 0.046$}& \multicolumn{2}{c}{$0.225 \pm 0.024$}& \multicolumn{2}{c}{$0.089 \pm 0.033$} \\
\multicolumn{3}{c}{DRCFR}& \multicolumn{2}{c}{$0.789 \pm 0.091$}& \multicolumn{2}{c}{$0.261 \pm 0.036$}& \multicolumn{2}{c}{$2.560\pm 0.210$}& \multicolumn{2}{c}{$0.332 \pm 0.053$}& \multicolumn{2}{c}{$0.235 \pm 0.015$}& \multicolumn{2}{c}{$0.119 \pm 0.045$} \\
\multicolumn{3}{c}{DeR-CFR}& \multicolumn{2}{c}{$0.529 \pm 0.068$}& \multicolumn{2}{c}{$0.147 \pm 0.022$}& \multicolumn{2}{c}{$1.857 \pm 0.109$}& \multicolumn{2}{c}{$0.261 \pm 0.045$}& \multicolumn{2}{c}{$0.208 \pm 0.062$}& \multicolumn{2}{c}{$0.093 \pm 0.032$} \\
\multicolumn{3}{c}{Our Method}& \multicolumn{2}{c}{$\bf{0.411} \pm 0.054$}& \multicolumn{2}{c}{$\bf{0.128} \pm 0.031$}& \multicolumn{2}{c}{$\bf{1.168} \pm 0.107$}& \multicolumn{2}{c}{$\bf{0.178} \pm 0.039$}& \multicolumn{2}{c}{$\bf{0.190} \pm 0.054$}& \multicolumn{2}{c}{$\bf{0.067} \pm 0.042$} \\
 \bottomrule
\end{tabular}
}
\end{center}
\end{table*}

\begin{table*}[t]
\footnotesize
\renewcommand{\arraystretch}{1.1}
\setlength{\tabcolsep}{8pt}
\caption{Ablation studies for different components of DDRN-CFR.}
\label{ablation-results}
\begin{center}
\scriptsize
\resizebox{ 0.91\textwidth}{!}{
\begin{tabular}{cccccccccccccccccccccccccc}
\toprule
\multicolumn{3}{c}{\multirow{2}{*}{HD}}& \multicolumn{3}{c}{\multirow{2}{*}{MMoE}}& \multicolumn{3}{c}{\multirow{2}{*}{MEMA}}& \multicolumn{3}{c}{\multirow{2}{*}{LOR}}& \multicolumn{3}{c}{\multirow{2}{*}{IL}}& \multicolumn{3}{c}{\multirow{2}{*}{ISW}}&  \multicolumn{4}{c}{\bf Within-sample}& \multicolumn{4}{c}{\bf Out-of-sample} \\
\cmidrule(l){19-22} \cmidrule(l){23-26} 
\multicolumn{3}{c}{}& \multicolumn{3}{c}{}& \multicolumn{3}{c}{}& \multicolumn{3}{c}{}& \multicolumn{3}{c}{}& \multicolumn{3}{c}{}& \multicolumn{2}{c}{PEHE}& \multicolumn{2}{c}{$\epsilon_{\mathrm{ATE}}$}& \multicolumn{2}{c}{PEHE}& \multicolumn{2}{c}{$\epsilon_{\mathrm{ATE}}$}\\
\midrule[0.5pt]
\multicolumn{3}{c}{\checkmark}& \multicolumn{3}{c}{}& \multicolumn{3}{c}{}& \multicolumn{3}{c}{\checkmark}& \multicolumn{3}{c}{\checkmark} & \multicolumn{3}{c}{\checkmark} & \multicolumn{2}{c}{$1.643 \pm 0.104$}& \multicolumn{2}{c}{$0.194 \pm 0.044$}& \multicolumn{2}{c}{$1.857 \pm 0.109$}& \multicolumn{2}{c}{$0.261 \pm 0.045$} \\
\multicolumn{3}{c}{}& \multicolumn{3}{c}{\checkmark}& \multicolumn{3}{c}{}& \multicolumn{3}{c}{\checkmark}& \multicolumn{3}{c}{\checkmark} & \multicolumn{3}{c}{\checkmark} & \multicolumn{2}{c}{$1.237 \pm 0.098$}& \multicolumn{2}{c}{$0.176 \pm 0.037$}& \multicolumn{2}{c}{$1.310 \pm 0.105$}& \multicolumn{2}{c}{$0.204 \pm 0.041$} \\
\multicolumn{3}{c}{}& \multicolumn{3}{c}{}& \multicolumn{3}{c}{\checkmark}& \multicolumn{3}{c}{}& \multicolumn{3}{c}{\checkmark}& \multicolumn{3}{c}{\checkmark} & \multicolumn{2}{c}{$1.275 \pm 0.102$}& \multicolumn{2}{c}{$0.153 \pm 0.039$}& \multicolumn{2}{c}{$1.324 \pm 0.092$}& \multicolumn{2}{c}{$0.207 \pm 0.047$} \\
\multicolumn{3}{c}{}& \multicolumn{3}{c}{}& \multicolumn{3}{c}{\checkmark}& \multicolumn{3}{c}{\checkmark}& \multicolumn{3}{c}{}& \multicolumn{3}{c}{\checkmark}  & \multicolumn{2}{c}{$1.163 \pm 0.094$}& \multicolumn{2}{c}{$0.145 \pm 0.038$}& \multicolumn{2}{c}{$1.249 \pm 0.104$}& \multicolumn{2}{c}{$0.181 \pm 0.040$} \\
\multicolumn{3}{c}{}& \multicolumn{3}{c}{}& \multicolumn{3}{c}{\checkmark}& \multicolumn{3}{c}{\checkmark}& \multicolumn{3}{c}{\checkmark}& \multicolumn{3}{c}{}  & \multicolumn{2}{c}{$1.211 \pm 0.087$}& \multicolumn{2}{c}{$0.157 \pm 0.042$}& \multicolumn{2}{c}{$1.292 \pm 0.116$}& \multicolumn{2}{c}{$0.190 \pm 0.044$} \\
\multicolumn{3}{c}{}& \multicolumn{3}{c}{}& \multicolumn{3}{c}{\checkmark}& \multicolumn{3}{c}{\checkmark}& \multicolumn{3}{c}{\checkmark}& \multicolumn{3}{c}{\checkmark} & \multicolumn{2}{c}{$\bf{1.074} \pm 0.095$}& \multicolumn{2}{c}{$\bf{0.126} \pm 0.036$}& \multicolumn{2}{c}{$\bf{1.168} \pm 0.107$}& \multicolumn{2}{c}{$\bf{0.178} \pm 0.039$} \\
\bottomrule
\end{tabular}
}
\end{center}
\end{table*}

\subsection{Experiment Results}


We report the results, including the mean and standard deviation (std) of treatment effect over 100 replications on IHDP, 10 replications on ACIC 2016 and Jobs datasets in Table~\ref{exp-results}. In the interest of fairness, we directly referred to the results from~\cite{AnpengWu2022} for the performance of the baseline models.
As we can observe,  our method is significantly better than the compared methods such as CFR-ISW~\cite{Hassanpour2019}, DRCFR~\cite{Hassanpour&Greiner2020} and DeR-CFR~\cite{AnpengWu2022} on a wide range of datasets. Moreover, we also verified the model effect on the real-world industrial datasets, such as homepage message pop-up dataset, through offline evaluation and online A/B test.  
Table~\ref{offline-results} and Figure \ref{fig:auuc_ddrn} illustrate the offline AUUC, Qini coefficient and uplift curves, which demonstrate that DDRN-CFR performs better than other methods on the real-world production dataset.

%

\begin{table*}[ht]
    \centering
    \begin{minipage}{0.49\textwidth}
        \begin{center}
        \footnotesize
        \renewcommand{\arraystretch}{1.15}
        \setlength{\tabcolsep}{12pt}
        \caption{The offline evaluation results of AUUC and Qini coefficient (mean±std) for the real-world production dataset.}
         \label{offline-results}
        \resizebox{0.99\textwidth}{!}{
	\begin{tabular}{cccccccccccc}
	\toprule   
	\multicolumn{4}{c}{\bf Methods}& \multicolumn{4}{c}{\bf AUUC}& \multicolumn{4}{c}{\bf Qini} \\
	\midrule[0.5pt]
	\multicolumn{4}{c}{DRCFR}& \multicolumn{4}{c}{$0.0309 \pm{0.0027}$}& \multicolumn{4}{c}{$0.0412\pm{0.0032}$} \\
	\multicolumn{4}{c}{DeR-CFR}& \multicolumn{4}{c}{$0.0346 \pm{0.0023}$}& \multicolumn{4}{c}{$0.0453 \pm{0.0034}$} \\
	\multicolumn{4}{c}{Our Method}& \multicolumn{4}{c}{$\bf{0.0372} \pm{0.0021}$}& \multicolumn{4}{c}{$\bf{0.0476} \pm{0.0029}$} \\
	\bottomrule
	\end{tabular}}  
	\end{center}
	\end{minipage}
    \hspace{0.01\textwidth}
    \begin{minipage}{0.49\textwidth}
        \begin{center}
        \footnotesize
        \renewcommand{\arraystretch}{1.15}
        \setlength{\tabcolsep}{2pt}
        \caption{Online A/B test evaluation results.\\ \hfill}
         \label{online-results}
        \resizebox{0.99\textwidth}{!}{
        \begin{tabular}{ccccccccccc}
	\toprule
	\multicolumn{3}{c}{\bf Metrics}& \multicolumn{4}{c}{\bf CTR}& \multicolumn{4}{c}{\bf DAC} \\
	\cmidrule(r){1-3} \cmidrule(lr){4-7} \cmidrule(l){8-11} 
	\multicolumn{3}{c}{Methods}& \multicolumn{2}{c}{Daily Sales}& \multicolumn{2}{c}{Major Promotion}& \multicolumn{2}{c}{Daily Sales}& \multicolumn{2}{c}{Major Promotion} \\
	\midrule[0.5pt]
	\multicolumn{3}{c}{DeR-CFR}& \multicolumn{2}{c}{+2.04\%}& \multicolumn{2}{c}{+1.29\%}& \multicolumn{2}{c}{+1.06\%}& \multicolumn{2}{c}{+0.47\%} \\
	\multicolumn{3}{c}{Our Method}& \multicolumn{2}{c}{\bf{+3.78\%}}& \multicolumn{2}{c}{\bf{+2.40\%}}& \multicolumn{2}{c}{\bf{+1.79\%}}& \multicolumn{2}{c}{\bf{+1.35\%}} \\
	\bottomrule
	\end{tabular} } 
	\end{center}
    \end{minipage}
\end{table*}

We designed an online A/B test to further evaluate performance by calculating the Click-Through Rate (CTR) and Daily Active Customers (DAC) of the population. Table~\ref{online-results} reports online DAC and CTR results of different models during 7 days experiment period. 
Specifically, compared to the current online baseline, the CTR increased by 3.78\% during the daily sales period and by 2.40\% during the major promotion period, while the DAC increased by 1.79\% during the daily sales period and by 1.35\% during the major promotion period. This again demonstrates the improvement of DDRN-CFR on ITE estimation.

\subsection{Ablation Study}

We also conduct an ablation study on ACIC 2016 dataset to investigate the contributions of different components in DDRN-CFR. Table~\ref{ablation-results} reports the results of the ablation experiments,
where HD denotes the traditional  \textbf{H}ard \textbf{D}ecomposition approach, which involves using three networks at the representation layer that are either completely independent
or merely shared-bottom. MMoE stands for \textbf{M}ulti-gate \textbf{M}ixture \textbf{o}f \textbf{E}xperts, MEMA refers to \textbf{M}ixture of \textbf{E}xperts with \textbf{M}ulti-head \textbf{A}ttention, LOR means \textbf{L}inear \textbf{O}rthogonal \textbf{R}egularizer, ISW and IL represent \textbf{I}mportant 
\textbf{S}ampling 
\textbf{W}eighting and \textbf{I}mbalance \textbf{L}oss,
respectively. According to Table~\ref{ablation-results}, we can draw the following conclusions: First of all, our proposed latent space soft separation method substantially outperforms hard decomposing methods. Secondly,  each component of DDRN-CFR is necessary, since lacking any one of them would damage the effect of ITE estimation on ACIC 2016.

Look beyond the outcome to grasp the essence. Compared to the traditional HD method, both MMoE and MEMA can substantially improve model performance, the core reason being the enhanced ability to extract information.
Considering that MEMA can leverage the powerful multi-head self-attention mechanism to fully account for the connections among different expert networks and adaptively learn the weights between them,
it therefore performs better than MMoE. When attempting to remove LOR module, we find that it incurs a certain degree of performance penalty, 
indicating that obtaining precise disentangled representations is highly valuable. 
Additionally, we must point out that the IL and ISW modules have brought positive performance gains by balancing confounders and mitigating selection bias.

\begin{figure*}[htbp]
\centering
\includegraphics[width=0.88\textwidth,keepaspectratio]{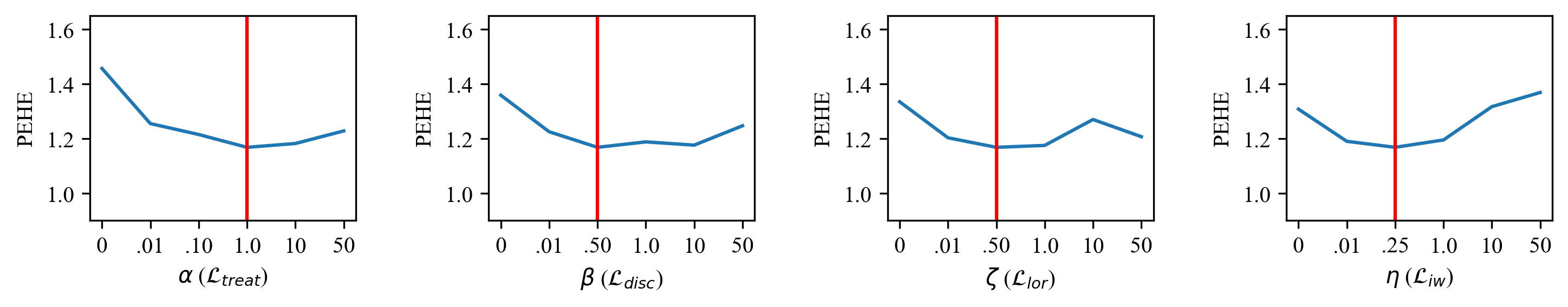}
\caption{Sensitivity analysis of the hyperparameters \{$\alpha, \beta, \zeta, \eta$\} on the ACIC 2016 dataset. The blue line shows the accuracy of the these parameters across different value ranges. The red line indicates the best parameters for the setting.}
\label{fig:hypers}
\end{figure*}

\subsection{Hyperparameters Analysis}

\begin{table}[b]
\footnotesize
\renewcommand{\arraystretch}{1.1}
\setlength{\tabcolsep}{10pt}
\caption{Hyperparameters and ranges}
\label{hypers}
\begin{center}
\scriptsize
\resizebox{0.48\textwidth}{!}{
\begin{tabular}{cccccccc}
\toprule
\multicolumn{4}{c}{\bf Hyperparameter}& \multicolumn{4}{c}{\bf Range}\\
\midrule[0.5pt]
\multicolumn{4}{c}{hidden state dimension}& \multicolumn{4}{c}{\{100, 200\}} \\
\multicolumn{4}{c}{mini-batch size}& \multicolumn{4}{c}{\{256, 512, 1024\}} \\
\multicolumn{4}{c}{the number of heads}& \multicolumn{4}{c}{\{2, 4, 6, 8\}} \\
\multicolumn{4}{c}{the output dimension of experts}& \multicolumn{4}{c}{\{100, 200, 300, 400\}} \\
\multicolumn{4}{c}{the number of layers in the experts}& \multicolumn{4}{c}{\{1, 2, 3\}} \\
\multicolumn{4}{c}{the number of layers in the towers}& \multicolumn{4}{c}{\{1, 2, 3\}} \\
\multicolumn{4}{c}{batch/layer normalization}& \multicolumn{4}{c}{\{False, True\}} \\
\bottomrule
\end{tabular}
}
\end{center}
\end{table}

Considering the complexity of the objective function with multiple terms, we conducted a grid search over the hyperparameters $\alpha$, $\beta$, $\zeta$, 
and $\eta$ within the value space $\{0.01, 0.10, 0.25, 0.50, 1.00, 10, 50\}$ to investigate the impact of each term on the accuracy of the potential outcomes. 
The experimental results shown in Figure \ref{fig:hypers} demonstrate that the performance of DDRN-CFR is primarily influenced by changes in the hyperparameter $\alpha$,
highlighting the necessity of accurately decomposing treatment factors.
While $\beta$ and $\eta$ may not significantly influence accuracy, they are essential conditions for balancing confounders and eliminating selection bias.
The remaining $\zeta$ will guarantee that three independent latent factors $\displaystyle \{ \Gamma, \Delta, \Upsilon \}$ are obtained through soft separation.
 
Furthermore, we use Adam as the optimizer, select GELU and ELU as the nonlinear activation functions for MEMA and task-specific towers respectively, and conduct approximately 1000 iterations to train the DDRN. Simultaneously, to mitigate overfitting, we typically set the regularization coefficient, also known as weight decay, within the range of
[1e-5, 1e-3].
Following hyperparameters analysis, we recommend setting hidden state dimension to 200, $\alpha$ to 1.0, $\eta$ to 0.25, $\beta$ and $\zeta$ to 0.50 each, and limiting the number of heads to no more than 4. In summary, our experimental results across various datasets reveal that hyperparameter selection is not a particularly complex challenge.
For details on our additional hyperparameters search space, see Table \ref{hypers} .
%

\section{Conclusions And Future Work}
In this paper, we focus on the problem of estimating individual treatment effect in observational studies. In treatment effect estimation, unmeasured or entangled confounders 
would easily fool the estimator to draw erroneous conclusions. Whereas most previous solutions primarily focus on either confounder balancing or importance sampling, 
they often ignore the criticality of confounder identification. Although
some promising algorithms have been proposed for confounder selection or disentanglement, they can not guarantee whether the confounder factors can be accurately identified.
Moreover, the model network architecture has low fault tolerance. Therefore, we propose a deep disentangled representation network for causal inference that 
incorporates mixture of experts with multi-head attention and a linear orthogonal regularizer to softly decompose the pre-treatment variables.
Experiments on a variety of benchmark datasets for causal inference demonstrate that our algorithm outperforms the state-of-the-art methods substantially.
However, this method is not suitable for the commonly encountered isomorphic multi-interventions in e-commerce settings.
In the future, we plan to further explore solutions for multi-treatment and continuous value scenarios, and attempt to apply them in a wider range of contexts.

\bibliographystyle{ACM-Reference-Format}
\bibliography{sigir}


\begin{thebibliography}{40}


\ifx \showCODEN    \undefined \def \showCODEN     #1{\unskip}     \fi
\ifx \showDOI      \undefined \def \showDOI       #1{#1}\fi
\ifx \showISBNx    \undefined \def \showISBNx     #1{\unskip}     \fi
\ifx \showISBNxiii \undefined \def \showISBNxiii  #1{\unskip}     \fi
\ifx \showISSN     \undefined \def \showISSN      #1{\unskip}     \fi
\ifx \showLCCN     \undefined \def \showLCCN      #1{\unskip}     \fi
\ifx \shownote     \undefined \def \shownote      #1{#1}          \fi
\ifx \showarticletitle \undefined \def \showarticletitle #1{#1}   \fi
\ifx \showURL      \undefined \def \showURL       {\relax}        \fi
\providecommand\bibfield[2]{#2}
\providecommand\bibinfo[2]{#2}
\providecommand\natexlab[1]{#1}
\providecommand\showeprint[2][]{arXiv:#2}

\bibitem[{A. Smith} and {E. Todd}(2005)]%
        {Smith&Todd2005}
\bibfield{author}{\bibinfo{person}{Jeffrey {A. Smith}} {and}
  \bibinfo{person}{Petra {E. Todd}}.} \bibinfo{year}{2005}\natexlab{}.
\newblock \showarticletitle{Does matching overcome LaLonde's critique of
  nonexperimental estimators?}
\newblock \bibinfo{journal}{\emph{Journal of Econometrics}}
  \bibinfo{volume}{125}, \bibinfo{number}{1} (\bibinfo{year}{2005}),
  \bibinfo{pages}{305--353}.
\newblock
\showISSN{0304-4076}
\urldef\tempurl%
\url{https://doi.org/10.1016/j.jeconom.2004.04.011}
\showDOI{\tempurl}
\newblock
\shownote{Experimental and non-experimental evaluation of economic policy and
  models}.


\bibitem[Arjovsky et~al\mbox{.}(2017)]%
        {arjovsky2017wasserstein}
\bibfield{author}{\bibinfo{person}{Martin Arjovsky}, \bibinfo{person}{Soumith
  Chintala}, {and} \bibinfo{person}{L{\'e}on Bottou}.}
  \bibinfo{year}{2017}\natexlab{}.
\newblock \bibinfo{title}{Wasserstein GAN}.
\newblock
\newblock
\showeprint[arxiv]{1701.07875}~[stat.ML]


\bibitem[Austin(2011)]%
        {austin2011introduction}
\bibfield{author}{\bibinfo{person}{Peter~C Austin}.}
  \bibinfo{year}{2011}\natexlab{}.
\newblock \showarticletitle{An introduction to propensity score methods for
  reducing the effects of confounding in observational studies}.
\newblock \bibinfo{journal}{\emph{Multivariate behavioral research}}
  \bibinfo{volume}{46}, \bibinfo{number}{3} (\bibinfo{year}{2011}),
  \bibinfo{pages}{399--424}.
\newblock


\bibitem[Bansal et~al\mbox{.}(2018)]%
        {orthreg2018}
\bibfield{author}{\bibinfo{person}{Nitin Bansal}, \bibinfo{person}{Xiaohan
  Chen}, {and} \bibinfo{person}{Zhangyang Wang}.}
  \bibinfo{year}{2018}\natexlab{}.
\newblock \showarticletitle{Can we gain more from orthogonality regularizations
  in training deep CNNs?}. In \bibinfo{booktitle}{\emph{Proceedings of the 32nd
  International Conference on Neural Information Processing Systems}}
  (Montr\'{e}al, Canada) \emph{(\bibinfo{series}{NIPS'18})}.
  \bibinfo{publisher}{Curran Associates Inc.}, \bibinfo{address}{Red Hook, NY,
  USA}, \bibinfo{pages}{4266--4276}.
\newblock


\bibitem[Cheng et~al\mbox{.}(2022)]%
        {cheng2022learning}
\bibfield{author}{\bibinfo{person}{Mingyuan Cheng}, \bibinfo{person}{Xinru
  Liao}, \bibinfo{person}{Quan Liu}, \bibinfo{person}{Bin Ma},
  \bibinfo{person}{Jian Xu}, {and} \bibinfo{person}{Bo Zheng}.}
  \bibinfo{year}{2022}\natexlab{}.
\newblock \showarticletitle{Learning disentangled representations for
  counterfactual regression via mutual information minimization}. In
  \bibinfo{booktitle}{\emph{Proceedings of the 45th International ACM SIGIR
  Conference on Research and Development in Information Retrieval}}.
  \bibinfo{pages}{1802--1806}.
\newblock


\bibitem[Devriendt et~al\mbox{.}(2022)]%
        {learn2rankum}
\bibfield{author}{\bibinfo{person}{Floris Devriendt}, \bibinfo{person}{Jente
  Van~Belle}, \bibinfo{person}{Tias Guns}, {and} \bibinfo{person}{Wouter
  Verbeke}.} \bibinfo{year}{2022}\natexlab{}.
\newblock \showarticletitle{Learning to Rank for Uplift Modeling}.
\newblock \bibinfo{journal}{\emph{IEEE Transactions on Knowledge and Data
  Engineering}} \bibinfo{volume}{34}, \bibinfo{number}{10}
  (\bibinfo{year}{2022}), \bibinfo{pages}{4888--4904}.
\newblock
\urldef\tempurl%
\url{https://doi.org/10.1109/TKDE.2020.3048510}
\showDOI{\tempurl}


\bibitem[Diemert et~al\mbox{.}(2018)]%
        {hal02515860}
\bibfield{author}{\bibinfo{person}{Eustache Diemert}, \bibinfo{person}{Artem
  Betlei}, \bibinfo{person}{Christophe Renaudin}, {and}
  \bibinfo{person}{Massih-Reza Amini}.} \bibinfo{year}{2018}\natexlab{}.
\newblock \showarticletitle{{A Large Scale Benchmark for Uplift Modeling}}. In
  \bibinfo{booktitle}{\emph{{Proceedings of the AdKDD and TargetAd Workshop,
  KDD, London,United Kingdom, August, 20, 2018}}}. \bibinfo{publisher}{{ACM}},
  \bibinfo{address}{London, United Kingdom}.
\newblock


\bibitem[Dorie et~al\mbox{.}(2019)]%
        {VincentDorie2019}
\bibfield{author}{\bibinfo{person}{Vincent Dorie}, \bibinfo{person}{Jennifer
  Hill}, \bibinfo{person}{Uri Shalit}, \bibinfo{person}{Marc Scott}, {and}
  \bibinfo{person}{Dan Cervone}.} \bibinfo{year}{2019}\natexlab{}.
\newblock \showarticletitle{{Automated versus Do-It-Yourself Methods for Causal
  Inference: Lessons Learned from a Data Analysis Competition}}.
\newblock \bibinfo{journal}{\emph{Statist. Sci.}} \bibinfo{volume}{34},
  \bibinfo{number}{1} (\bibinfo{year}{2019}), \bibinfo{pages}{43 -- 68}.
\newblock
\urldef\tempurl%
\url{https://doi.org/10.1214/18-STS667}
\showDOI{\tempurl}


\bibitem[Gretton et~al\mbox{.}(2012)]%
        {gretton12a}
\bibfield{author}{\bibinfo{person}{Arthur Gretton}, \bibinfo{person}{Karsten~M.
  Borgwardt}, \bibinfo{person}{Malte~J. Rasch}, \bibinfo{person}{Bernhard
  Sch{{\"o}}lkopf}, {and} \bibinfo{person}{Alexander Smola}.}
  \bibinfo{year}{2012}\natexlab{}.
\newblock \showarticletitle{A Kernel Two-Sample Test}.
\newblock \bibinfo{journal}{\emph{Journal of Machine Learning Research}}
  \bibinfo{volume}{13}, \bibinfo{number}{25} (\bibinfo{year}{2012}),
  \bibinfo{pages}{723--773}.
\newblock
\urldef\tempurl%
\url{http://jmlr.org/papers/v13/gretton12a.html}
\showURL{%
\tempurl}


\bibitem[Hainmueller(2012)]%
        {Hainmueller_2012}
\bibfield{author}{\bibinfo{person}{Jens Hainmueller}.}
  \bibinfo{year}{2012}\natexlab{}.
\newblock \showarticletitle{Entropy Balancing for Causal Effects: A
  Multivariate Reweighting Method to Produce Balanced Samples in Observational
  Studies}.
\newblock \bibinfo{journal}{\emph{Political Analysis}} \bibinfo{volume}{20},
  \bibinfo{number}{1} (\bibinfo{year}{2012}), \bibinfo{pages}{25--46}.
\newblock
\urldef\tempurl%
\url{https://doi.org/10.1093/pan/mpr025}
\showDOI{\tempurl}


\bibitem[Hassanpour and Greiner(2019)]%
        {Hassanpour2019}
\bibfield{author}{\bibinfo{person}{Negar Hassanpour} {and}
  \bibinfo{person}{Russell Greiner}.} \bibinfo{year}{2019}\natexlab{}.
\newblock \showarticletitle{CounterFactual Regression with Importance Sampling
  Weights}. In \bibinfo{booktitle}{\emph{Proceedings of the Twenty-Eighth
  International Joint Conference on Artificial Intelligence, {IJCAI-19}}}.
  \bibinfo{publisher}{International Joint Conferences on Artificial
  Intelligence Organization}, \bibinfo{pages}{5880--5887}.
\newblock
\urldef\tempurl%
\url{https://doi.org/10.24963/ijcai.2019/815}
\showDOI{\tempurl}


\bibitem[Hassanpour and Greiner(2020)]%
        {Hassanpour&Greiner2020}
\bibfield{author}{\bibinfo{person}{Negar Hassanpour} {and}
  \bibinfo{person}{Russell Greiner}.} \bibinfo{year}{2020}\natexlab{}.
\newblock \showarticletitle{Learning Disentangled Representations for
  CounterFactual Regression}. In \bibinfo{booktitle}{\emph{International
  Conference on Learning Representations}}.
\newblock
\urldef\tempurl%
\url{https://openreview.net/forum?id=HkxBJT4YvB}
\showURL{%
\tempurl}


\bibitem[Hernan and Robins(2020)]%
        {hernan2023causal}
\bibfield{author}{\bibinfo{person}{M.A. Hernan} {and} \bibinfo{person}{J.M.
  Robins}.} \bibinfo{year}{2020}\natexlab{}.
\newblock \bibinfo{booktitle}{\emph{Causal Inference: What If}}.
\newblock \bibinfo{publisher}{CRC Press}.
\newblock
\showISBNx{9781420076165}
\showLCCN{2022050839}
\urldef\tempurl%
\url{https://books.google.com/books?id=_KnHIAAACAAJ}
\showURL{%
\tempurl}


\bibitem[Hill(2011)]%
        {hill21011}
\bibfield{author}{\bibinfo{person}{Jennifer Hill}.}
  \bibinfo{year}{2011}\natexlab{}.
\newblock \showarticletitle{Bayesian Nonparametric Modeling for Causal
  Inference}.
\newblock \bibinfo{journal}{\emph{Journal of Computational and Graphical
  Statistics}}  \bibinfo{volume}{20} (\bibinfo{date}{03} \bibinfo{year}{2011}),
  \bibinfo{pages}{217--240}.
\newblock
\urldef\tempurl%
\url{https://doi.org/10.1198/jcgs.2010.08162}
\showDOI{\tempurl}


\bibitem[Imbens and Rubin(2015)]%
        {imbens2015causal}
\bibfield{author}{\bibinfo{person}{Guido~W. Imbens} {and}
  \bibinfo{person}{Donald~B. Rubin}.} \bibinfo{year}{2015}\natexlab{}.
\newblock \bibinfo{booktitle}{\emph{Causal Inference for Statistics, Social,
  and Biomedical Sciences: An Introduction}}.
\newblock \bibinfo{publisher}{Cambridge University Press}.
\newblock
\showISBNx{9781139025751}


\bibitem[Joffe et~al\mbox{.}(2004)]%
        {Joffe2004}
\bibfield{author}{\bibinfo{person}{Marshall Joffe}, \bibinfo{person}{Thomas
  Have}, \bibinfo{person}{Harold Feldman}, {and} \bibinfo{person}{Stephen
  Kimmel}.} \bibinfo{year}{2004}\natexlab{}.
\newblock \showarticletitle{Model Selection, Confounder Control, and Marginal
  Structural Models: Review and New Applications}.
\newblock \bibinfo{journal}{\emph{The American Statistician}}
  \bibinfo{volume}{58} (\bibinfo{date}{02} \bibinfo{year}{2004}),
  \bibinfo{pages}{272--279}.
\newblock


\bibitem[Johansson et~al\mbox{.}(2016)]%
        {Johansson&Shalit2016}
\bibfield{author}{\bibinfo{person}{Fredrik~D. Johansson}, \bibinfo{person}{Uri
  Shalit}, {and} \bibinfo{person}{David Sontag}.}
  \bibinfo{year}{2016}\natexlab{}.
\newblock \showarticletitle{Learning Representations for Counterfactual
  Inference}. In \bibinfo{booktitle}{\emph{Proceedings of the 33rd
  International Conference on International Conference on Machine Learning -
  Volume 48}} (New York, NY, USA) \emph{(\bibinfo{series}{ICML'16})}.
  \bibinfo{publisher}{JMLR.org}, \bibinfo{pages}{3020--3029}.
\newblock


\bibitem[Kohavi and Longbotham(2011)]%
        {Kohavi2011}
\bibfield{author}{\bibinfo{person}{Ron Kohavi} {and} \bibinfo{person}{Roger
  Longbotham}.} \bibinfo{year}{2011}\natexlab{}.
\newblock \showarticletitle{Unexpected results in online controlled
  experiments}.
\newblock \bibinfo{journal}{\emph{SIGKDD Explor. Newsl.}} \bibinfo{volume}{12},
  \bibinfo{number}{2} (\bibinfo{date}{mar} \bibinfo{year}{2011}),
  \bibinfo{pages}{31--35}.
\newblock
\showISSN{1931-0145}
\urldef\tempurl%
\url{https://doi.org/10.1145/1964897.1964905}
\showDOI{\tempurl}


\bibitem[Kuang et~al\mbox{.}(2019)]%
        {Kuang2019}
\bibfield{author}{\bibinfo{person}{Kun Kuang}, \bibinfo{person}{Peng Cui},
  \bibinfo{person}{Bo Li}, \bibinfo{person}{Meng Jiang},
  \bibinfo{person}{Yashen Wang}, \bibinfo{person}{Fei Wu}, {and}
  \bibinfo{person}{Shiqiang Yang}.} \bibinfo{year}{2019}\natexlab{}.
\newblock \showarticletitle{Treatment effect estimation via differentiated
  confounder balancing and regression}.
\newblock \bibinfo{journal}{\emph{ACM Transactions on Knowledge Discovery from
  Data (TKDD)}} \bibinfo{volume}{14}, \bibinfo{number}{1}
  (\bibinfo{year}{2019}), \bibinfo{pages}{1--25}.
\newblock


\bibitem[Kuang et~al\mbox{.}(2017)]%
        {KuangKun2017}
\bibfield{author}{\bibinfo{person}{Kun Kuang}, \bibinfo{person}{Peng Cui},
  \bibinfo{person}{Bo Li}, \bibinfo{person}{Meng Jiang},
  \bibinfo{person}{Shiqiang Yang}, {and} \bibinfo{person}{Fei Wang}.}
  \bibinfo{year}{2017}\natexlab{}.
\newblock \showarticletitle{Treatment Effect Estimation with Data-Driven
  Variable Decomposition}.
\newblock \bibinfo{journal}{\emph{Proceedings of the AAAI Conference on
  Artificial Intelligence}} \bibinfo{volume}{31}, \bibinfo{number}{1}
  (\bibinfo{date}{Feb.} \bibinfo{year}{2017}).
\newblock
\urldef\tempurl%
\url{https://doi.org/10.1609/aaai.v31i1.10480}
\showDOI{\tempurl}


\bibitem[LaLonde(1986)]%
        {LaLonde1986}
\bibfield{author}{\bibinfo{person}{Robert~J. LaLonde}.}
  \bibinfo{year}{1986}\natexlab{}.
\newblock \showarticletitle{Evaluating the Econometric Evaluations of Training
  Programs with Experimental Data}.
\newblock \bibinfo{journal}{\emph{American Economic Review}}
  \bibinfo{volume}{76} (\bibinfo{year}{1986}), \bibinfo{pages}{604--20}.
\newblock


\bibitem[Louizos et~al\mbox{.}(2017)]%
        {louizos2017causal}
\bibfield{author}{\bibinfo{person}{Christos Louizos}, \bibinfo{person}{Uri
  Shalit}, \bibinfo{person}{Joris Mooij}, \bibinfo{person}{David Sontag},
  \bibinfo{person}{Richard Zemel}, {and} \bibinfo{person}{Max Welling}.}
  \bibinfo{year}{2017}\natexlab{}.
\newblock \bibinfo{title}{Causal Effect Inference with Deep Latent-Variable
  Models}.
\newblock
\newblock
\showeprint[arxiv]{1705.08821}~[stat.ML]


\bibitem[Lunceford and Davidian(2004)]%
        {Lunceford2004}
\bibfield{author}{\bibinfo{person}{Jared~K. Lunceford} {and}
  \bibinfo{person}{Marie Davidian}.} \bibinfo{year}{2004}\natexlab{}.
\newblock \showarticletitle{Stratification and weighting via the propensity
  score in estimation of causal treatment effects: a comparative study}.
\newblock \bibinfo{journal}{\emph{Statistics in Medicine}}
  \bibinfo{volume}{23} (\bibinfo{year}{2004}).
\newblock
\urldef\tempurl%
\url{https://api.semanticscholar.org/CorpusID:11912618}
\showURL{%
\tempurl}


\bibitem[Ma et~al\mbox{.}(2018)]%
        {MMoE}
\bibfield{author}{\bibinfo{person}{Jiaqi Ma}, \bibinfo{person}{Zhe Zhao},
  \bibinfo{person}{Xinyang Yi}, \bibinfo{person}{Jilin Chen},
  \bibinfo{person}{Lichan Hong}, {and} \bibinfo{person}{Ed~H. Chi}.}
  \bibinfo{year}{2018}\natexlab{}.
\newblock \showarticletitle{Modeling Task Relationships in Multi-task Learning
  with Multi-gate Mixture-of-Experts}. In \bibinfo{booktitle}{\emph{Proceedings
  of the 24th ACM SIGKDD International Conference on Knowledge Discovery \&
  Data Mining}} (London, United Kingdom) \emph{(\bibinfo{series}{KDD '18})}.
  \bibinfo{publisher}{Association for Computing Machinery},
  \bibinfo{address}{New York, NY, USA}, \bibinfo{pages}{1930--1939}.
\newblock
\showISBNx{9781450355520}
\urldef\tempurl%
\url{https://doi.org/10.1145/3219819.3220007}
\showDOI{\tempurl}


\bibitem[Pearl(2009a)]%
        {10.1214/09-SS057}
\bibfield{author}{\bibinfo{person}{Judea Pearl}.}
  \bibinfo{year}{2009}\natexlab{a}.
\newblock \showarticletitle{{Causal inference in statistics: An overview}}.
\newblock \bibinfo{journal}{\emph{Statistics Surveys}} \bibinfo{volume}{3},
  \bibinfo{number}{none} (\bibinfo{year}{2009}), \bibinfo{pages}{96 -- 146}.
\newblock
\urldef\tempurl%
\url{https://doi.org/10.1214/09-SS057}
\showDOI{\tempurl}


\bibitem[Pearl(2009b)]%
        {causality}
\bibfield{author}{\bibinfo{person}{Judea Pearl}.}
  \bibinfo{year}{2009}\natexlab{b}.
\newblock \bibinfo{booktitle}{\emph{Causality: Models, Reasoning and Inference}
  (\bibinfo{edition}{2nd} ed.)}.
\newblock \bibinfo{publisher}{Cambridge University Press}.
\newblock


\bibitem[Pearl et~al\mbox{.}(2016)]%
        {pearl2016causal}
\bibfield{author}{\bibinfo{person}{J. Pearl}, \bibinfo{person}{M. Glymour},
  {and} \bibinfo{person}{N.P. Jewell}.} \bibinfo{year}{2016}\natexlab{}.
\newblock \bibinfo{booktitle}{\emph{Causal Inference in Statistics: A Primer}}.
\newblock \bibinfo{publisher}{Wiley}.
\newblock
\showISBNx{9781119186847}
\showLCCN{2015037219}
\urldef\tempurl%
\url{https://books.google.com/books?id=L3G-CgAAQBAJ}
\showURL{%
\tempurl}


\bibitem[Rosenbaum and Rubin(1983)]%
        {RosenbaumandRubin1983}
\bibfield{author}{\bibinfo{person}{Paul~R. Rosenbaum} {and}
  \bibinfo{person}{Donald~B. Rubin}.} \bibinfo{year}{1983}\natexlab{}.
\newblock \showarticletitle{The central role of the propensity score in
  observational studies for causal effects}.
\newblock \bibinfo{journal}{\emph{Biometrika}} \bibinfo{volume}{70},
  \bibinfo{number}{1} (\bibinfo{date}{04} \bibinfo{year}{1983}),
  \bibinfo{pages}{41--55}.
\newblock
\showISSN{0006-3444}
\urldef\tempurl%
\url{https://doi.org/10.1093/biomet/70.1.41}
\showDOI{\tempurl}


\bibitem[Rubin(1974)]%
        {RubinDonaldB1974Eceo}
\bibfield{author}{\bibinfo{person}{Donald~B Rubin}.}
  \bibinfo{year}{1974}\natexlab{}.
\newblock \showarticletitle{Estimating causal effects of treatments in
  randomized and nonrandomized studies}.
\newblock \bibinfo{journal}{\emph{Journal of educational psychology}}
  \bibinfo{volume}{66}, \bibinfo{number}{5} (\bibinfo{year}{1974}),
  \bibinfo{pages}{688--701}.
\newblock
\showISSN{0022-0663}


\bibitem[Rubin(1997)]%
        {rubin1997}
\bibfield{author}{\bibinfo{person}{Donald~B Rubin}.}
  \bibinfo{year}{1997}\natexlab{}.
\newblock \showarticletitle{Estimating Causal Effects from Large Data Sets
  Using Propensity Scores}.
\newblock \bibinfo{journal}{\emph{Annals of Internal Medicine}}
  \bibinfo{volume}{127} (\bibinfo{year}{1997}), \bibinfo{pages}{757--763}.
\newblock


\bibitem[Schafer and Kang(2009)]%
        {Schafer&Kang2009}
\bibfield{author}{\bibinfo{person}{Joseph Schafer} {and}
  \bibinfo{person}{Joseph Kang}.} \bibinfo{year}{2009}\natexlab{}.
\newblock \showarticletitle{Average Causal Effects From Nonrandomized Studies:
  A Practical Guide and Simulated Example}.
\newblock \bibinfo{journal}{\emph{Psychological methods}}  \bibinfo{volume}{13}
  (\bibinfo{date}{01} \bibinfo{year}{2009}), \bibinfo{pages}{279--313}.
\newblock
\urldef\tempurl%
\url{https://doi.org/10.1037/a0014268}
\showDOI{\tempurl}


\bibitem[Shalit et~al\mbox{.}(2017)]%
        {shalit2017estimating}
\bibfield{author}{\bibinfo{person}{Uri Shalit}, \bibinfo{person}{Fredrik~D.
  Johansson}, {and} \bibinfo{person}{David Sontag}.}
  \bibinfo{year}{2017}\natexlab{}.
\newblock \showarticletitle{Estimating individual treatment effect:
  generalization bounds and algorithms}. In
  \bibinfo{booktitle}{\emph{Proceedings of the 34th International Conference on
  Machine Learning}} \emph{(\bibinfo{series}{ICML'17})}.
  \bibinfo{address}{Sydney, NSW, Australia}.
\newblock


\bibitem[Shi et~al\mbox{.}(2019)]%
        {shi2019adapting}
\bibfield{author}{\bibinfo{person}{Claudia Shi}, \bibinfo{person}{David~M.
  Blei}, {and} \bibinfo{person}{Victor Veitch}.}
  \bibinfo{year}{2019}\natexlab{}.
\newblock \bibinfo{title}{Adapting Neural Networks for the Estimation of
  Treatment Effects}.
\newblock
\newblock
\showeprint[arxiv]{1906.02120}~[stat.ML]


\bibitem[Shimodaira(2000)]%
        {SHIMODAIRA2000227}
\bibfield{author}{\bibinfo{person}{Hidetoshi Shimodaira}.}
  \bibinfo{year}{2000}\natexlab{}.
\newblock \showarticletitle{Improving predictive inference under covariate
  shift by weighting the log-likelihood function}.
\newblock \bibinfo{journal}{\emph{Journal of Statistical Planning and
  Inference}} \bibinfo{volume}{90}, \bibinfo{number}{2} (\bibinfo{year}{2000}),
  \bibinfo{pages}{227--244}.
\newblock
\showISSN{0378-3758}
\urldef\tempurl%
\url{https://doi.org/10.1016/S0378-3758(00)00115-4}
\showDOI{\tempurl}


\bibitem[Tuff{\'e}ry(2011)]%
        {upliftcurves}
\bibfield{author}{\bibinfo{person}{St{\'e}phane Tuff{\'e}ry}.}
  \bibinfo{year}{2011}\natexlab{}.
\newblock \bibinfo{booktitle}{\emph{Data Mining and Statistics for Decision
  Making}}.
\newblock \bibinfo{publisher}{Wiley}.
\newblock


\bibitem[Vaswani et~al\mbox{.}(2023)]%
        {vaswani2023attention}
\bibfield{author}{\bibinfo{person}{Ashish Vaswani}, \bibinfo{person}{Noam
  Shazeer}, \bibinfo{person}{Niki Parmar}, \bibinfo{person}{Jakob Uszkoreit},
  \bibinfo{person}{Llion Jones}, \bibinfo{person}{Aidan~N. Gomez},
  \bibinfo{person}{Lukasz Kaiser}, {and} \bibinfo{person}{Illia Polosukhin}.}
  \bibinfo{year}{2023}\natexlab{}.
\newblock \bibinfo{title}{Attention Is All You Need}.
\newblock
\newblock
\showeprint[arxiv]{1706.03762}~[cs.CL]


\bibitem[Wu et~al\mbox{.}(2022)]%
        {AnpengWu2022}
\bibfield{author}{\bibinfo{person}{Anpeng Wu}, \bibinfo{person}{Junkun Yuan},
  \bibinfo{person}{Kun Kuang}, \bibinfo{person}{Bo Li}, \bibinfo{person}{Runze
  Wu}, \bibinfo{person}{Qiang Zhu}, \bibinfo{person}{Yueting Zhuang}, {and}
  \bibinfo{person}{Fei Wu}.} \bibinfo{year}{2022}\natexlab{}.
\newblock \showarticletitle{Learning decomposed representations for treatment
  effect estimation}.
\newblock \bibinfo{journal}{\emph{IEEE Transactions on Knowledge and Data
  Engineering}} \bibinfo{volume}{35}, \bibinfo{number}{5}
  (\bibinfo{year}{2022}), \bibinfo{pages}{4989--5001}.
\newblock


\bibitem[Zhang et~al\mbox{.}(2019)]%
        {zhanglu2019}
\bibfield{author}{\bibinfo{person}{Lu Zhang}, \bibinfo{person}{Yongkai Wu},
  {and} \bibinfo{person}{Xintao Wu}.} \bibinfo{year}{2019}\natexlab{}.
\newblock \showarticletitle{Causal Modeling-Based Discrimination Discovery and
  Removal: Criteria, Bounds, and Algorithms}.
\newblock \bibinfo{journal}{\emph{IEEE Transactions on Knowledge and Data
  Engineering}} \bibinfo{volume}{31}, \bibinfo{number}{11}
  (\bibinfo{year}{2019}), \bibinfo{pages}{2035--2050}.
\newblock
\urldef\tempurl%
\url{https://doi.org/10.1109/TKDE.2018.2872988}
\showDOI{\tempurl}


\bibitem[Zhang et~al\mbox{.}(2021)]%
        {zhang2021treatment}
\bibfield{author}{\bibinfo{person}{Weijia Zhang}, \bibinfo{person}{Lin Liu},
  {and} \bibinfo{person}{Jiuyong Li}.} \bibinfo{year}{2021}\natexlab{}.
\newblock \showarticletitle{Treatment effect estimation with disentangled
  latent factors}. In \bibinfo{booktitle}{\emph{Proceedings of the AAAI
  Conference on Artificial Intelligence}}, Vol.~\bibinfo{volume}{35}.
  \bibinfo{pages}{10923--10930}.
\newblock


\bibitem[Zubizarreta(2015)]%
        {zubi2015}
\bibfield{author}{\bibinfo{person}{Jos{\'e} Zubizarreta}.}
  \bibinfo{year}{2015}\natexlab{}.
\newblock \showarticletitle{Stable Weights that Balance Covariates for
  Estimation With Incomplete Outcome Data}.
\newblock \bibinfo{journal}{\emph{J. Amer. Statist. Assoc.}}
  \bibinfo{volume}{110} (\bibinfo{date}{04} \bibinfo{year}{2015}),
  \bibinfo{pages}{0--0}.
\newblock
\urldef\tempurl%
\url{https://doi.org/10.1080/01621459.2015.1023805}
\showDOI{\tempurl}


\end{thebibliography}

%
%
%
%
%
%
%
%

\end{document}